\documentclass{article}

\usepackage[preprint,nonatbib]{neurips_2020}

\usepackage[utf8]{inputenc} 
\usepackage[T1]{fontenc}    
\usepackage[colorlinks,linkcolor=red,anchorcolor=blue,citecolor=deepgreen]{hyperref}       
\usepackage[pdftex]{graphicx}
\usepackage{url}            
\usepackage{booktabs}       
\usepackage{amsfonts}       
\usepackage{amssymb}
\usepackage{amsmath}
\usepackage{nicefrac}       
\usepackage{microtype}      
\usepackage{graphicx}
\usepackage{color}

\usepackage{multicol}
\usepackage{multirow}
\usepackage{colortbl}
\definecolor{gray}{RGB}{130,130,130} 
\definecolor{tels}{RGB}{250,250,250}
\definecolor{newton}{RGB}{242,242,242}
\definecolor{deepgreen}{RGB}{39,158,41}

\newtheorem{theorem}{Theorem}

\title{GNN-XML: Graph Neural Networks for Extreme Multi-label Text Classification}

\author{%
 Daoming Zong \\
  East China Normal University\\
  \texttt{ecnuzdm@gmail.com} \\
  \And
  Shiliang Sun \\
  East China Normal University\\
  \texttt{slsun@cs.ecnu.edu.cn} \\
  }

\begin{document}

  \maketitle

  \begin{abstract}
  Extreme multi-label text classification (XMTC) aims to tag a text instance with the most relevant subset of labels from an extremely large label set. XMTC has attracted much recent attention due to massive label sets yielded by modern applications, such as news annotation and product recommendation. The main challenges of XMTC are the data \textit{scalability} and \textit{sparsity}, thereby leading to two issues: i) the intractability to scale to the extreme label setting, ii) the presence of long-tailed label distribution, implying that a large fraction of labels have few positive training instances. To overcome these problems, we propose GNN-XML, a scalable graph neural network framework tailored for XMTC problems. Specifically, we exploit label correlations via mining their co-occurrence patterns and build a label graph based on the correlation matrix. We then conduct the attributed graph clustering by performing graph convolution with a low-pass graph filter to jointly model label dependencies and label features, which induces semantic label clusters. We further propose a bilateral-branch graph isomorphism network to decouple representation learning and classifier learning for better modeling \textit{tail labels}. Experimental results on multiple benchmark datasets show that GNN-XML significantly outperforms state-of-the-art methods while maintaining comparable prediction efficiency and model size. 
  \end{abstract}

  \section{Introduction}
  Extreme multi-label text classification (XMTC) aims to assign to each given text its most relevant subset of labels from an extremely large label set, where the number of labels could reach millions or more. Recently, XMTC has attracted increasing attention due to the rapid growth of web-scale data in many applications, such as news annotation, web page tagging~\cite{partalas2015lshtc}, product categorization for e-commerce, Bing’s dynamic search advertising~\cite{prabhu2014fastxml,partalas2015lshtc,prabhu2018parabel}, to name a few.

  XMTC poses great computational challenges for developing effective and efficient classifiers due to the extreme number of labels. The huge label space has raised research challenges such as data \textit{sparsity} and \textit{scalability}. Figure~\ref{long-tail} demonstrates an example of the long-tailed label distribution in the Wiki10-31K dataset~\cite{zubiaga2012enhancing}, where a large fraction of labels have very few positive training instances, namely \textit{tail labels}. A variety of approaches have been proposed to address the challenges of \textit{scalability} and label \textit{sparsity} in the XMTC problems. Despite One-vs-All (OVA) approaches~\cite{babbar2017dismec,babbar2019data} that generally achieve strong performance, they suffer from scalability issues. Following a \textit{divide-and-conquer} paradigm, tree-based methods~\cite{prabhu2014fastxml,jain2016extreme} scale to large label sets in XMTC by hierarchically partitioning the instance or label space. However, they are prone to error propagation in the tree cascade and tend to perform inferiorly on \textit{tail labels} prediction~\cite{prabhu2018parabel}. Label-embedding methods~\cite{bi2013efficient,cisse2013robust} reduce the label space by projecting label vectors onto a low-dimensional space based on the assumption that the label matrix is low-rank. Nevertheless, such assumption is violated in real world applications owing to the presence of \textit{tail labels}~\cite{bhatia2015sparse,xu2016robust,tagami2017annexml}. We refer the readers to Sup.~{\color{red}{A}} for details of these approaches.


Deep learning models benefiting from their powerful text representation capabilities have also been explored for the XMTC problems~\cite{liu2017deep,attentionxml,chang2019x}. XML-CNN~\cite{liu2017deep} first applies a dynamic max-pooling scheme and a family of CNN models to learn text representations. AttentionXML~\cite{attentionxml} uses the attentional bidirectional long short-term memory (BiLSTM) networks to extract embeddings from raw text inputs. X-BERT~\cite{chang2019x} first incorporates the bidirectional encoder representations from transformers (BERT~\cite{devlin2018bert}) for XMTC. Despite the empirical success of these methods, there are still some concerns required to be addressed. For example, the CNN-based models cannot capture the most relevant parts of the input to each label, the RNN-based methods fail to model long-term dependencies due to vanishing gradients. X-BERT simply uses truncation techniques to process the text with a length larger than 512, which may discard the crucial information for classification. As shown in Fig.~\ref{long-tail}, 83.8\% of the training instances have more than 512 words. In addition, the difficulty of scaling to the extreme label space remains in deep learning methods as the output layer scales linearly with the product of label size and feature dimension~\cite{chang2019x}. Data \textit{sparsity} among tail labels also makes the instance volume much lower than that required by deep models to reach peak performance~\cite{khandagale2019bonsai}.

In this paper, we propose GNN-XML, a scalable graph neural network framework tailored for the XMTC problem. In a nutshell, we achieve the \textit{scalability} by partitioning the enormous number of labels into smaller number of clusters with a novel label partitioning method. By parsing each input document into a keyword graph, we utilize the \textit{graph isomorphism network} (GIN) to fully explore the deep semantic context information and learn robust representations for matching input text to a small set of label clusters. We further propose the bilateral-balancing branch learning to gradually place more emphasis on \textit{tail labels}. Our main contributions are summarized as follows:
\begin{figure}[!t]
\begin{tabular}{@{}c@{}}
\includegraphics[width=1.0\textwidth]{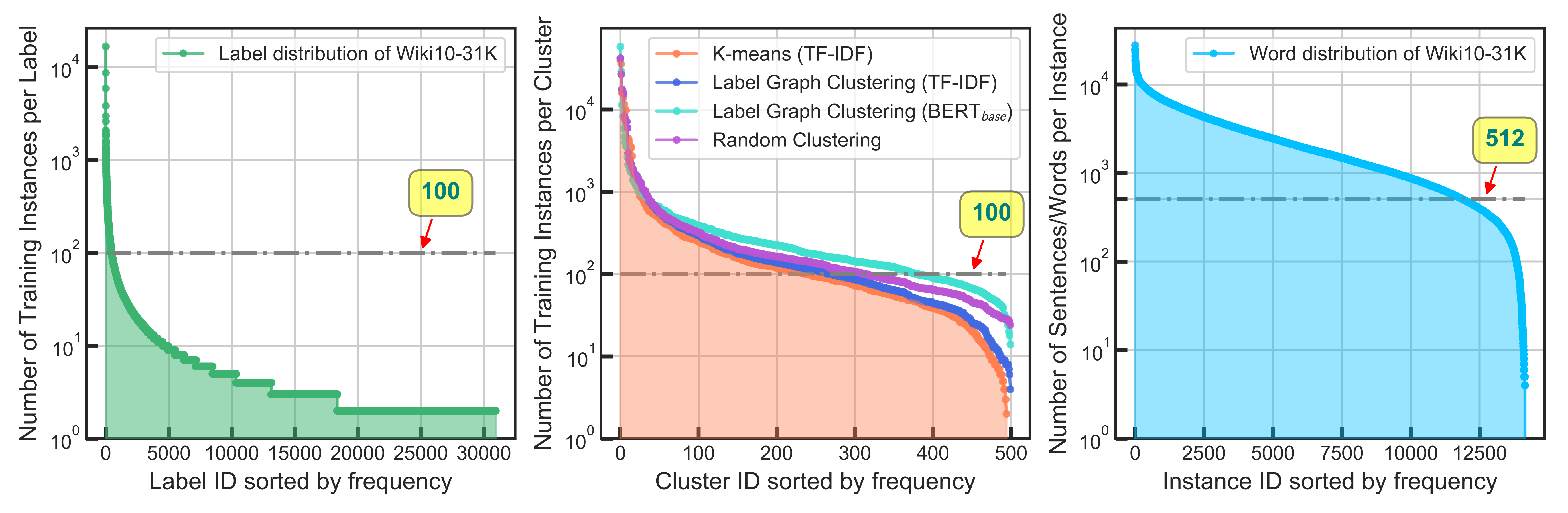}
\end{tabular}
\caption{\textit{Left:} An illustration of a long-tailed  (power-law) label distribution of Wiki10-31K. Note that only 1.51\% of the labels have more than 100 training instances. \textit{Middle:} Distributions of the clusters varying with different label clustering approaches. 75.8\% of the clusters have more than 100 training instances by our label graph clustering (BERT$_{\text{base}}$), which largely mitigates the \textit{sparsity} issue. \textit{Right:} Word distribution of Wiki10-31K. Notably, 83.8\% of the instances have more than 512 words.}
\label{long-tail}
\end{figure}

\begin{itemize}
  \item We exploit the correlations between labels via mining their co-occurrence patterns within the dataset and build a label graph based on the correlation matrix. We conduct the attributed graph clustering with a low-pass graph filter to jointly model node/label connectivity and node/label features, which induces dependency-aware and semantic-aware label clusters.

  \item The KeyGraph building schema allows flexible integration of various pre-trained language representation models, such as BERT and its variants, which endows GNN-XML with powerful text representational capabilities.

  \item We propose an end-to-end bilateral-branch graph isomorphism network to decouple representation learning and classifier learning for \textit{tail labels}. Ablation studies are performed to demonstrate the significance and effectiveness of reasoning and mining \textit{tail labels}.
\end{itemize}

\section{Preliminaries} 
\label{sec:preliminaries}
\paragraph{Problem Setup and Notations} Formally, let $\mathcal{X}=\mathbb{R}^{d}$ denote the $d$-dimensional input space and $\mathcal{Y} =\{0,1\}^{L}$ denote the label space with $L$ unique labels. The task of multi-label classification is to learn a function $f: \mathcal{X}\rightarrow 2^{\mathcal{Y}}$ that maps an instance $\mathbf{x}\in \mathcal{X}$ to its targets $\mathbf{y}=[y_{1}, y_{2}, \ldots , y_{L}] \in \mathcal{Y}$. 
\paragraph{Graph Neural Networks} 
\label{par:graph_neural_networks}
GNNs use the graph structure (node connectivity) and node features $x_{v}$ to learn a representation vector of a node, $h_{v}$, or the entire graph, $h_{G}$. Most existing GNNs follow a neighborhood aggregation strategy, where the representation of a node is iteratively updated by aggregating representations of its neighbors. After $k$ iterations of aggregation, a node's representation captures the structure information within its $k$-hop neighborhood. The $k$-th layer of a GNN is,
\begin{equation}
\begin{aligned}
a_{v}^{(k)}=\text{AGGREGATE}^{(k)}(\{h_{u}^{(k-1)}:u\in \mathcal{N}(v)\}),~h_{v}^{(k)}=\text{COMBINE}^{(k)}(h_{v}^{(k-1)},a_{v}^{(k)}),
\end{aligned}
\end{equation}
where $h_{v}^{(k)}$ is the feature vector of node $v$ at the $k$-th iteration/layer. Initially, $h_{v}^{(0)}=x_{v}$. $\mathcal{N}(v)$ is a set of nodes adjacent to $v$. AGGREGATE essentially assembles a node’s neighborhood information while COMBINE controls message passing across different hops/layers. The choice of AGGREGATE and COMBINE in GNNs is crucial, where a number of architectures for AGGREGATE and COMBINE have been proposed~\cite{kipf2016semi,hamilton2017inductive,xu2018representation}. Node embeddings learned by GNNs can be directly used for tasks like node classification and link prediction. For graph classification tasks, a READOUT~\cite{xu2018powerful,zhang2018end,lee2019self} function is that, given embeddings of individual nodes, produces the embedding of the entire graph. Unlike the success of other deep models that intensively rely on deep layers, GNNs with few layers could achieve convincing performance~\cite{xu2018powerful,yao2019graph}, which implies a relative small model size.


\section{The Proposed Model} 
\label{sec:the_proposed_method}

\begin{figure}[!t]
\begin{tabular}{@{}c@{}}
\includegraphics[width=1.0\textwidth]{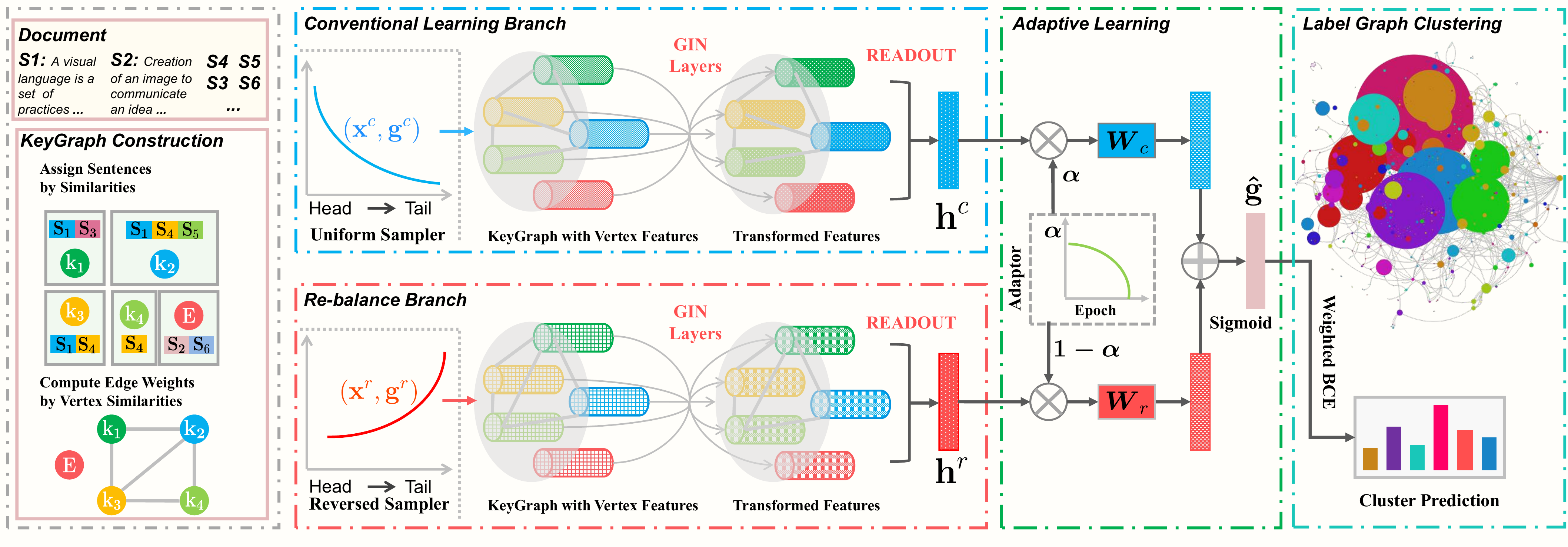}
\end{tabular}
\caption{Illustration of GNN-XML pipeline. First, construct two KeyGraphs for a pair of text inputs respectively and send them to the bilateral-branch GIN, where i) the conventional learning branch takes inputs from a uniform sampler, accounting for learning universal patterns of the original distributions; ii) the re-balance branch takes inputs from a reversed sampler tailored for \textit{tail labels}. Then, two branches are aggregated by the adaptive learning strategy to calculate the probabilities of assigning the pair of text inputs to their relevant clusters induced by the label graph clustering.}
\label{overview}
\end{figure}
\paragraph{Probabilistic Mixture Motivations} 
\label{par:mixture_models_for_XMTC}
We formulate our framework for XMTC in a probabilistic mixture perspective. Formally, given an instance $\mathbf{x}$ and its label $\mathbf{y}$, a mixture model introduces a multinomial latent variable $z \in \{1,\ldots,K\}$ and decomposes the marginal likelihood as:
\begin{equation}
p(\mathbf{y}|\mathbf{x})=\sum_{z=1}^{K}p(\mathbf{y},z|\mathbf{x})=\sum_{z=1}^{K}p(z|\mathbf{x};\Phi)p(\mathbf{y}|z,\mathbf{x};\Theta),
\end{equation}
where $z$ is a hidden categorical indicator variable, such that $z=k$ indicates that the instance is assigned to the $k$-th \textit{component}. The prior $p(z|\mathbf{x};\Phi)$ assigns each instance to its relevant \textit{components} probabilistically. The likelihood $p(\mathbf{y}|z,\mathbf{x};\Theta)$ estimates the probabilities of getting label $y_{\ell}$ from the $k$-th \textit{cluster} for an instance $\mathbf{x}$, and can be instantiated by any binary classifier which provides probability estimations. Assume after label clustering (each \textit{cluster} is regarded as a \textit{component} here), we have $K$ clusters of labels, $\{\mathcal{C}_{k}\}_{k=1,\ldots,K}$, where $\mathcal{C}_{k}$ denotes a subset of the label indices of the $k$-th \textit{cluster}, i.e., $\mathcal{C}_{k}\in [L]:=\{1,\ldots,L\}$. For labels that do not fall into the ${k}$-th \textit{cluster}, we impose that $p(y_{\ell}=1|z=k,\mathbf{x};\Theta)=0$ if $\ell \not \in \mathcal{C}_{k}$. The $p(\mathbf{y}|z,\mathbf{x};\Theta)$ is thus factorized as a product of independent conditional binary label distributions:
\begin{equation}
\begin{aligned}
p(\mathbf{y}|z=k,\mathbf{x};\Theta)= \prod_{\ell \in L} p(y_{\ell}|z=k,\mathbf{x};\Theta) \approx\prod_{\ell \in \mathcal{C}_{k}} p(y_{\ell}|z=k,\mathbf{x};\Theta).
\end{aligned}
\label{classifier}
\end{equation}
As shown in Fig.~\ref{overview}, GNN-XML is mainly composed of two sub-modules, i.e., the graph matching model which is responsible for matching the input text to a set of relevant label clusters, and the label clustering model that accounts for partitioning the extreme number of labels into $K$ \textit{clusters}.

\subsection{Graph Matching Model} 
\label{sub:graph_matching_}
\paragraph{KeyGraph Construction} 
Given an input text $\mathbf{x}_{i}$, we first perform named entity recognition and keyword extraction by TextRank~\cite{mihalcea2004textrank}. Inspired by~\cite{liu2019matching}, we then build a keyword co-occurrence graph, named KeyGraph, based on the set of found keywords. Each keyword is a vertex in the KeyGraph. If vertices $v_{i}$ and $v_{j}$ share at least one sentence, we add an edge $e_{ij}$ between them. The edge weight $w_{i,j}$ is calculated by the number of shared sentences. Note that one sentence can be associated with multiple keywords, which implies connections between different topics. Sentences that do not match any keywords in the document will be attached to an \textit{empty} vertex that does not contain any keyword.

\paragraph{Vertex Representation} For each sentence, we use BERT-base~\cite{devlin2018bert} embeddings as its representation. Note that we have also explored other recent pre-trained language representation models for comparison, including RoBERTa~\cite{liu2019roberta}, XLNET~\cite{yang2019xlnet} and ALBERT~\cite{lan2019albert}. For each vertex $v$, we take the sum of the embeddings of the sentences attached to it as its initial representation $h_{v}^{(0)}$.

\paragraph{Bilateral-Branch Graph Isomorphism Network} Among many powerful GNNs, we adopt the graph isomorphism network~\cite{xu2018powerful} (GIN) for text representations of two branches considering its outstanding expressive capacity and model simplicity. The propagation rule of GIN is defined as:
\begin{equation}
\begin{aligned}
h_{v}^{(k)}=\text{MLP}^{(k)}\left((1+\epsilon^{(k)})\cdot h_{v}^{(k-1)}+\sum\nolimits_{u\in \mathcal{N}(v)}h_{u}^{(k-1)}\right),
\end{aligned}
\label{gin}
\end{equation}
where $\epsilon$ is a learnable parameter or a fixed scalar. With the increasing number of iterations/depths, node representations become more refined and global, which is key to achieving good discriminative power. Yet, features from earlier iterations may sometimes generalize better~\cite{xu2018representation}. Therefore, we use all structural information by concatenating graph representations across all iterations of GIN, and instantiate READOUT with the summing of all node features from the same iterations as follows:
\begin{equation}
\begin{aligned}
h_{G}=\text{CONCAT}(\text{SUM}(\{h_{v}^{(k)}|v\in G\})|k=0,1,\ldots,K).
\end{aligned}
\end{equation}

As illustrated in Fig. \ref{overview}, our model consists of two branches accounting for representation learning and classifier learning, called \textit{conventional learning branch} and \textit{re-balance branch}, respectively. The input data for the conventional learning branch comes from a uniform sampler, where each sample in the training set is sampled only once with equal probability. The uniform sampler retains the original data distribution, so it is conducive to representation learning. While, the re-balance branch aims to alleviate severely long-tailed label distributions and improve the recognition capability on \textit{tail labels}, which is favorable for classifier learning. For the reversed sampler, the sampling possibility of each label is inversely proportional to its number of relevant instances, that is, the more instances in the label, the smaller the label sampling probability. Formally, let $n_{\ell}$ denote the number of relevant instances of a label ${\ell}$ and $n_{max}$ denote the maximum sample number of all labels. To construct the reversed sampler, we need: 1) calculate the sampling possibility of label $y_{\ell}$ according to the number of samples, $p_{\ell}={w_{\ell}}/{\sum_{i=1}^{L}}w_{i}$ where $w_{\ell}=n_{max}/n_{\ell}$; 2) randomly sample a label according to $p_{\ell}$; 3) uniformly pick up a sample from label ${\ell}$ with replacement. By repeating the uniform and reversed sampling procedures, pairwise training data of a mini-batch can be obtained.

\paragraph{Adaptive Learning} 
To avoid heavy over-fitting on \textit{tail labels} and insufficient training on \textit{head labels}, we use an adaptive learning strategy to reconcile the bilateral branch learning. Concretely, the balance factor, denoted by $\alpha$, shifts the learning attention between the bilateral branches by controlling both the weights for the features $\mathbf{h}^{c},\mathbf{h}^{r}\in \mathbb{R}^{D}$ learned from two branches and the cluster matching loss $\mathcal{L}$. The value of $\alpha$ is set to gradually decrease as the training epochs increases, such as the parabolic decay~\cite{zhou2019bbn} with $1-(t/t_{max})^{2}$, where $t$ denotes the current epoch and $t_{max}$ denotes the maximum training epochs. This suggests that the learning focus of our model gradually changes from universal patterns to tail data. Stick to the previous notations, the \textit{cluster} $\mathcal{C}_{k}$ is considered to be relevant to an instance $\mathbf{x}_{i}$ if the instance has positive label in $\mathcal{C}_{k}$, i.e., $y_{i\ell}=1, \exists \ell \in \mathcal{C}_{k}$. Let $\mathbf{g}_{i}=[g_{1},\ldots,g_{K}] \in \{0,1\}^{K}$ represent the ground truth indices of the relevant clusters for instance $\mathbf{x}_{i}$. During training, let $\hat{\mathbf{g}}$ denote the mixed matching probability for each \textit{cluster} of a pair of text inputs $(\mathbf{x}_{i}^{c},\mathbf{g}_{i}^{c})$ and $(\mathbf{x}^{r}_{i},\mathbf{g}_{i}^{r})$, which is given by 
\begin{equation}
\begin{aligned}
\hat{\mathbf{g}}_{i} = \sigma(\alpha{W}_{c}\mathbf{h}^{c}_{i}+(1-\alpha){W}_{r}\mathbf{h}^{r}_{i}),  
\end{aligned}
\end{equation}
where $W_{c} \in \mathbb{R}^{K\times D}$ and $W_{r}\in \mathbb{R}^{K\times D}$ are classifiers belonging to each branch, whose outputs will be integrated together by element-wise addition and activated by the sigmoid function $\sigma$. We then apply a weighted binary cross-entropy loss, denoted by $E(\cdot,\cdot)$, for optimization of the pairwise inputs:
\begin{equation}
\label{matching_loss}
\begin{aligned} 
\mathcal{L}=\alpha E(\hat{\mathbf{g}}_{i},\mathbf{g}^{c}_{i}) + (1-\alpha)E(\hat{\mathbf{g}}_{i},\mathbf{g}^{r}_{i}).
\end{aligned}
\end{equation}
During inference, the test samples are fed into both branches with fixed $\alpha=0.5$ as both branches are equally important. Then, the equally weighted features are fed to their corresponding classifiers to compute two predicted scores and both scores are aggregated by element-wise addition to return the predicated cluster indices. Once the label clusters are determined, the remain task is reduced to a small number of independent binary classification problems. See Sup. {\color{red}{D}} for algorithmic details.

\subsection{Label Clustering Model} 
\label{sub:graph_clustering_model}
\paragraph{Correlation Matrix} 
\label{par:correlation_matrix}
To build the correlation matrix, we first count the occurrence times of each label in the training set, denoted as the matrix ${N}\in \mathbb{R}^{L}$, where $L$ is the number of total unique labels. Then we count the occurrence times of all label pairs and obtain the label co-occurrence matrix ${M} \in \mathbb{R}^{L\times L}$. The conditional probability matrix ${P} \in \mathbb{R}^{L\times L}$ is thus given by ${P}_{ij}={{M}_{ij}}/{{N}_{i}}$, where ${P}_{ij}=p(y_{j}|y_{i})$ denotes the occurrence probability of label $y_{j}$ when label $y_{i}$ appears. To avoid the negative effect by the weakly correlated labels, we use a threshold $\rho$ to filter noisy edges and obtain a binary correlation matrix. However, the binary correlation matrix may lead to \textit{over-smoothing}~\cite{li2018deeper,chen2019multi}, which makes node features indistinguishable. Hence, we further resort to a re-weighted scheme:
\begin{equation}
{B}_{ij}=
\begin{cases}
0, &  \text{ if }{P}_{ij} < \rho \\ 
1, & \text{ if } {P}_{ij}\geq \rho
\end{cases}~~~~~~~~~
{A}_{ij}=
\begin{cases}
\tau/\sum\nolimits_{j\neq i}^{L}{B}_{ij}, & \text{ if } i \neq j \\ 
1-\tau, & \text{ if } i=j
\end{cases}
\end{equation}
where $\tau$ controls the edge weights assigned to a node itself (self-loop) and its neighbors. $B$ is the  binary correlation matrix, and ${A}$ is the final re-weighted correlation matrix. 

\paragraph{Label Embedding} 
\label{par:label_embeddings}
The label representation $\mathbf{z}_{\ell}$ is the sum embeddings of all relevant instances for the $l$-th label, i.e., $\mathbf{z}_{\ell} =\mathbf{v}_{\ell}/||\mathbf{v}_{\ell}||,\mathbf{v}_{\ell}=\sum\nolimits_{i:y_{i\ell}=1}\phi(\mathbf{x}_{i})~\text{for}~\ell =1,\ldots,L$, where $\phi$ denotes the embedding function, which can be the sparse TF-IDF features~\cite{jain2019slice,khandagale2019bonsai,chang2019x} or the BERT embeddings. 

\paragraph{Label Graph Clustering} 
\label{par:attributed_graph_clustering}
Denote $\mathcal{G}_{{label}}=(\mathcal{V},\mathcal{E},A,Z)$ as the constructed label graph, where $\mathcal{V}$ is a set of label nodes with $|\mathcal{V}|=L$, $\mathcal{E}$ is a set of edges, ${A}=[a_{ij}]\in \mathbb{R}^{L\times L}$ is the adjacency matrix and $Z=[\mathbf{z}_{\ell}]_{\ell=1}^{L} \in \mathbb{R}^{L\times D}$ is the initial feature (label embedding) matrix. Denote by $D$ the degree matrix of ${A}$ where $d_{i}=\sum_{j}a_{ij}$ is the degree of node $v_{i}$. The symmetrically normalized graph Laplacian is defined as $L_{\text{s}}=I-D^{-\frac{1}{2}}AD^{-\frac{1}{2}}$, where $I$ is the identity matrix. Based on the cluster assumption that nearby nodes are likely to be in the same cluster, performing graph convolution with a low-pass graph filter will make the clustering task easier~\cite{wang2019attributed,zhang2019attributed}. Note that, the first order graph convolution may not be adequate to achieve this, especially for large and sparse graphs~\cite{zhang2019attributed}, as it updates each node $v_{i}$ by the aggregation of its $1$-hop neighbors only, ignoring long-distance neighborhood relations. We thus capture global graph structures, i.e., label correlations, by $k$-order graph convolution. Similar to~\cite{zhang2019attributed}, we choose $G=(I-0.5L_{s})$ as the low-pass graph filter. The filtered feature matrix is given by
\begin{equation}
\begin{aligned}
\bar{Z}=GZ=(I-\frac{1}{2}L_{s})^{k}Z.
\end{aligned}
\label{graph_convolution}
\end{equation}
Our goal is to partition the nodes of the label graph $\mathcal{G}_{label}$ into $K$ clusters. For simplicity and scalability, we then apply the mini-batch $k$-means algorithm~\cite{sculley2010web} on the filtered feature matrix $\bar{Z}$.


\subsection{Complexity Analysis} 
\label{sub:time_complexity}

\begin{theorem}(Proof in Sup.~{\color{red}{C}}) Denote by $n_{e}$ the number of nonzero entries of the adjacency matrix $A$, $d$ the dimension of attributes/features, $k$ the $k$-order graph convolution, $K$ the number of induced clusters, $M$ the sample size and $T$ the number of iterations of the mini-batch $k$-means. Performing label graph clustering on $\mathcal{G}_{label}$ costs $\mathcal{O}(n_{e}kd+MTKd)$ in time, the training complexity of graph matching is $\mathcal{O}(NKd)$ and the prediction time complexity is approximately $\mathcal{O}(Kd+ (\frac{L}{K}\log L)d)$. 
\label{claim}
\end{theorem}
According to theorem~\ref{claim}, the time complexity is dominated by the label clustering model, which is proportional to $n_{e}$. We note that $n_{e} \propto L\log L$~\cite{jain2016extreme,prabhu2018parabel,jain2019slice}, so it may not scale up well for extreme large label graphs. Therefore, similar to FastGCN~\cite{chen2018fastgcn}, we further develop an efficient sampling scheme on top of the vanilla min-batch algorithm to reduce both time and memory complexities. Suppose $B$ is the mini-batch size, $S$ is the sampling size and $S \ll L$. Our sampling strategy cuts down the time complexity from $\mathcal{O}(n_{e}kd)$ to $\mathcal{O}(SBkd)$ and the memory complexity from $\mathcal{O}(Lkd)$ to $\mathcal{O}(SBkd)$. By controlling the size $B$ and $S$, GNN-XML can scale to the extreme label settings. Moreover, our label clustering model is non-parametric and does not need to train neural network parameters, thereby enjoying additional time efficiency. The detailed proof is cast to the Sup.~{\color{red}{C}}.


\section{Experiments} 
\label{sec:empirical_results}
\subsection{Datasets and Evaluation Metrics} 
\label{sec:datasets_and_Evaluation_metrics}


We evaluate our model on six widely used XMTC benchmark datasets, including three large-scale datasets: EURLex-4K~\cite{mencia2008efficient}, Wiki10-31K~\cite{zubiaga2012enhancing}, and AmazonCat-13K~\cite{mcauley2013hidden}; and three extreme-scale datasets: Wiki-500K, Amazon-670K~\cite{mcauley2013hidden}, and Amazon-3M~\cite{mcauley2015inferring}. The statistics of the benchmarks are summarized in Table \ref{datasets}. Following previous XMTC methods~\cite{jain2016extreme,yen2016pd,yen2017ppdsparse,babbar2017dismec,babbar2019data}, we use four metrics for comprehensive evaluation, including P@$k$, nDCG@$k$, PSP@$k$, and PSnDCG@$k$, which are well explained in~\cite{jain2016extreme}. Notably, PSP@$k$, and PSnDCG@$k$ are metrics designed for tail labels.



\begin{table}[!hbpt]
\label{datasets}
\setlength{\tabcolsep}{5.1mm}{
\scriptsize
\begin{tabular}{crrrrrr}
\toprule
Dataset & $N_{train}$ & $N_{test}$ &  $D$ & $L$ & $\overline{L}$ & $\hat{L}$    \\ \hline
EURLex-4K    &15,539     &3,809  &500        &3,993   &5.31   &25.73   \\
Wiki10-31K   &14,146     &6,616  &101,938    &30,938  &18.64  &8.52    \\
AmazonCat-13K&1,186,239  &306,782 &203,882   &13,330  &5.04   &448.57   \\
Wiki-500K    &1,779,881  &769,421 &2,381,304 &501,008 &4.75   &16.86  \\
Amazon-670K   &490,449    &153,025 &135,909   &670,091 &5.45   &3.99    \\
Amazon-3M    &1,717,899  &742,507 &337,067   &2,812,281 &36.04&22.02  \\ \bottomrule
\end{tabular}}
\caption{Dataset Statistics. $N_{train}$: training instances, $N_{test}$: test instances, $D$: feature dimension; $L$: total number of class labels, $\overline{L}$: average labels per instance, $\hat{L}$: average instances per label.}
\end{table}

\subsection{Comparing Methods and Implementation Details} 
\label{sub:comparing_methods_and_parameter_settings}
We compare the proposed GNN-XML to state-of-the-art XMTC methods including the embedding-based methods SLEEC~\cite{bhatia2015sparse} and  AnnexML~\cite{tagami2017annexml}; the feature space partitioning method FastXML~\cite{prabhu2014fastxml} and PfastreXML~\cite{jain2016extreme}, the label space partitioning method Parabel~\cite{prabhu2018parabel}, Bonsai~\cite{khandagale2019bonsai} and SLICE~\cite{jain2019slice}; the 1-vs-All methods PD-Sparse~\cite{yen2016pd}, PPDSparse~\cite{yen2017ppdsparse}, DiSMEC~\cite{babbar2017dismec} and ProXML~\cite{babbar2019data}; and three most representative deep learning methods XML-CNN~\cite{liu2017deep}, AttentionXML~\cite{attentionxml} and X-BERT~~\cite{chang2019x}. Since achieving the reproducible performance is required to parallelize with at least 100 cores, which is unavailable in our machine, we directly take the results of PDSparse~\cite{yen2017ppdsparse} and DiSMEC~\cite{babbar2017dismec} from the Extreme Repository\footnote{http://manikvarma.org/downloads/XC/XMLRepository.html}. The remaining methods are all conducted in our machine by running their released codes on the benchmark dataset partitions for fair comparison. The hyper-parameters for all the compared algorithms were set using fine grained validation on each data set so as to achieve the highest possible prediction score for each method. All the experiments are conducted on the NVIDIA DGX-1 server with eight Tesla P100 GPUs (16 GB memory each), dual 20-Core Intel Xeon E5-2698v4 (2.2 GHz) CPUs and 512 GB of RAM. We set a memory limit to be 100G following~\cite{yen2017ppdsparse}. 

Our model is implemented in PyTorch DistributedDataParallel~\footnote{https://pytorch.org/docs/stable/distributed.html} module with NCCL as the backend. The pipeline of training linear classifier for each label is also parallelized as Equ.~\ref{classifier} can be optimized independently for each label, which is instantiated by logistic regression and implemented via LIBLINEAR~\cite{fan2008liblinear}. The train/validate/test split is following the conventional setting for deep learning scenario~\cite{liu2017deep}. Following~\cite{devlin2018bert}, we perform sentence segmentation, use WordPiece embeddings~\cite{wu2016google} with a 30,000 token vocabulary and denote split word pieces with \#\#. We use the pre-trained BERT-base model~\cite{devlin2018bert} with a hidden size of 768, 12 Transformer blocks~\cite{vaswani2017attention}. The parameter $\alpha$ in Equ.~\ref{gin} is set to 0 and the orders $k$ in Equ.~\ref{graph_convolution} is tuned to 3. We train GNN-XML using the RAdam~\cite{liu2019variance} optimizer with its original settings and save the best model if the validation loss does not decrease for 10 epochs. The initial learning rate is 0.01 with a warm-up proportion as 0.1. See Sup.~{\color{red}{D}} for experiment details. 




\begin{table}[]
\setlength{\tabcolsep}{0.5mm}{
\scriptsize
\begin{tabular}{r|ccc|ccc|ccc|ccc|ccc|ccc}
\toprule
& \multicolumn{3}{c|}{\textbf{EURLex-4K}} & \multicolumn{3}{c|}{\textbf{Wiki10-31K}} & \multicolumn{3}{c|}{\textbf{AmazonCat-13K}} & \multicolumn{3}{c|}{\textbf{Wiki-500K}} & \multicolumn{3}{c|}{\textbf{Amazon-670K}} & \multicolumn{3}{c}{\textbf{Amazon-3M}} \\
\textbf{Metric} & P@1     & P@3     & P@5     & P@1      & P@3      & P@5      & P@1       & P@3       & P@5       & P@1      & P@3      & P@5     & P@1       & P@3      & P@5      & P@1      & P@3      & P@5     \\ \hline

\rowcolor{tels}
SLEEC~\cite{bhatia2015sparse}         &{{79.26}}         &{{64.30}}         &{{52.33}}         &{{85.88}}          &{{72.98}}          &{{62.70}}          &{{90.53}}           &{{76.33}}           &{{61.52}}           &{{48.20}}          &{{29.40}}          &{{21.20}}         &{{35.05}}           &{{31.25}}          &{{28.56}}          &{{--}}          &{{--}}          &{{--}}      \\ 

\rowcolor{newton}
AnnexML~\cite{tagami2017annexml} &{{79.66}}         &{{64.94}}         &{{53.32}}         &{{86.46}}          &{{74.28}}          &{{64.20}}          &{{93.54}}           &{{78.36}}           &{{63.30}}           &{{64.22}}          &{{43.15}}          &{{32.79}}         &{{42.09}}           &{{36.61}}          &{{32.75}}          &{{49.30}}          &{{45.55}}          &{{43.11}}      \\ \hline

\rowcolor{tels}
FastXML~\cite{prabhu2014fastxml} &{{71.36}}         &{{59.90}}         &{{50.39}}         &{{83.03}}          &{{67.47}}          &{{57.76}}          &{{93.11}}           &{{78.20}}           &{{63.41}}           &{{54.10}}          &{{35.50}}          &{{26.20}}         &{{36.99}}           &{{33.28}}          &{{30.53}}          &{{44.24}}          &{{40.83}}          &{{38.59}}   \\ 

\rowcolor{newton}
PfastreXML~\cite{jain2016extreme}  &{{75.45}}         &{{62.70}}         &{{52.51}}         &{{83.57}}          &{{68.61}}          &{{59.10}}          &{{91.75}}           &{{77.97}}           &{{63.68}}           &{{59.20}}          &{{40.30}}          &{{30.70}}         &{{39.46}}           &{{35.81 }}          &{{33.05}}          &{{43.83}}          &{{41.81}}          &{{40.09}}   \\ 

\rowcolor{tels}
Parabel~\cite{prabhu2018parabel} &{{81.73}}         &{{68.78}}         &{{57.44}}         &{{84.31}}          &{{72.57}}          &{{63.39}}          &{{93.03}}           &{{79.16}}           &{{64.52}}           &{{67.50}}          &{{48.70}}          &{{37.70}}         &{{44.89}}           &{{39.80}}          &{{36.00}}          &{{47.48}}          &{{44.65}}          &{{42.53}}       \\ 

\rowcolor{newton}
Bonsai~\cite{khandagale2019bonsai}  &{{83.00}}         &{{69.70}}         &{{58.40}}         &{{84.52}}          &{{73.76}}          &{{64.69}}          &{{92.98}}           &{{79.13}}           &{{64.46}}       &{{69.20}}  &{{49.80}}          &{{38.83}}         &{{45.58}}    &{{40.39}}   &{{36.60}}     &{{48.45}}          &{{45.65}}          &{{43.49}}   \\ \hline

\rowcolor{tels}
SLICE~\cite{jain2019slice}    &{{84.32}}         &{{71.32}}         &{{59.45}}         &{{86.61}}          &{{74.26}}          &{{64.48}}          &{{93.56}}           &{{79.20}}           &{{65.21}}           &{{72.94}}          &{{52.24}}          &{{41.42}}         &{{48.10}}           &{{44.00}}          &{{40.92}}          &{{48.28}}          &{{44.86}}          &{{43.24}}    \\ 

\rowcolor{newton}
PD-Sparse~\cite{yen2016pd}  &{{76.43}}         &{{60.37}}         &{{49.72}}         &{{83.12}}          &{{67.26}}          &{{58.21}}          &{{90.60}}           &{{75.14}}           &{{60.69}}           &{{--}}          &{{--}}          &{{--}}         &{{--}}           &{{--}}          &{{--}}          &{{--}}          &{{--}}          &{{--}}  \\ 

\rowcolor{tels}
PPDSparse~\cite{yen2017ppdsparse}    &{{\underline{83.83}}}         &{{\underline{70.72}}}         &{{\underline{59.21}}}         &{{85.99}}          &{{74.63}}          &{{64.38}}          &{{\underline{92.72}}}           &{{\underline{78.14}}}           &{{\underline{63.41}}}           &{{72.62}}          &{{51.79}}          &{{40.57}}         &{{\underline{45.32}}}           &{{\underline{40.37}}}          &{{\underline{36.92}}}           &{{48.41}}          &{{45.25}}          &{{43.16}}      \\

\rowcolor{newton}
DiSMEC~\cite{babbar2017dismec}    &{{\underline{82.40}}}         &{{\underline{68.50}}}         &{{\underline{57.70}}}         &{{\underline{85.20}}}          &{{\underline{74.60}}}          &{{\underline{65.90}}}          &{{\underline{93.40}}}           &{{\underline{79.10}}}           &{{\underline{64.10}}}           &{{\underline{70.20}}}          &{{\underline{50.60}}}          &{{\underline{39.70}}}         &{{\underline{44.70}}}           &{{\underline{39.70}}}          &{{\underline{36.10}}}          &{{\underline{47.34}}}          &{{\underline{44.96}}}          &{{\underline{42.80}}}  \\ 

\rowcolor{tels}
ProXML~\cite{babbar2019data}    &{{83.40}}         &{{70.90}}         &{{59.10}}         &{{83.68}}          &{{71.42}}          &{{63.13}}          &{{92.92}}           &{{78.61}}           &{{64.37}}           &{{69.00}}          &{{49.10}}          &{{38.80}}         &{{43.50}}           &{{38.70}}          &{{35.30}}          &{{45.17}}          &{{41.16}}          &{{40.13}}   \\ \hline

\rowcolor{newton}
XML-CNN~\cite{liu2017deep}  &{{76.38}}         &{{62.81}}         &{{51.41}}         &{{81.41}}          &{{66.23}}          &{{56.11}}          &{{93.26}}           &{{77.06}}           &{{61.40}}           &{{69.85}}          &{{49.28}}          &{{39.81}}         &{{35.39}}           &{{31.93}}          &{{29.32}}          &{{--}}          &{{--}}          &{{--}}  \\ 

\rowcolor{tels}
AttentionXML~\cite{attentionxml}    &{{87.12}}         &{{73.99}}         &{{61.92}}         &{{87.47}}          &{{78.48}}          &{{69.37}}          &{{95.92}}           &{{82.41}}           &{{67.31}}           &{{76.95}}          &{{58.42}}          &{{46.14}}     &{{47.58}}          &{{42.61}}          &{{38.92}}      &{{50.86}}           &{{48.04}}          &{{45.83}}             \\

\rowcolor{newton}
X-BERT~~\cite{chang2019x}      &{{87.83}}         &{{74.52}}         &{{62.24}}    &{{88.41}}          &{{79.32}}          &{{70.28}}          &{{95.17}}           &{{80.65}}           &{{65.19}}           &{{76.87}}          &{{56.73}}          &{{45.97}}         &{{48.98}}           &{{44.16}}          &{{41.43}}          &{{51.95}}          &{{49.46}}          &{{46.41}}    \\

GNN-XML    &\textbf{89.36}        &\textbf{76.14}       &\textbf{64.32}       &  \textbf{90.41}         &\textbf{81.13}          &\textbf{72.32}          &\textbf{97.67}           &\textbf{84.68}         &\textbf{69.95}          &\textbf{78.24}         &\textbf{59.21}       &\textbf{47.46}         &\textbf{51.12}          &\textbf{47.13}         &\textbf{43.28}          &\textbf{52.53}         &\textbf{49.67}      &\textbf{47.31}    \\ \bottomrule

\textbf{Metric} & S@1     & S@3     & S@5     & S@1      & S@3      & S@5      & S@1       & S@3       & S@5       & S@1      & S@3      & S@5     & S@1       & S@3      & S@5      & S@1      & S@3      & S@5     \\ \hline

\rowcolor{tels}
SLEEC~\cite{bhatia2015sparse}         &{{34.25}}         &{{39.83}}         &{{42.76}}         &{{11.14}}          &{{11.86}}          &{{12.40}}          &{{46.75}}           &{{58.46}}           &{{65.96}}           &{{21.10}}          &{{21.00}}          &{{20.80}}         &{{20.62}}           &{{23.32}}          &{{25.98}}          &{{--}}          &{{--}}          &{{--}}      \\ 

\rowcolor{newton}
AnnexML~\cite{tagami2017annexml} &{{33.88}}         &{{40.29}}         &{{43.69}}         &{{11.86}}          &{{12.75}}          &{{13.57}}          &{{51.02}}           &{{65.57}}           &{{70.13}}           &{{23.90}}          &{{28.14}}          &{{31.24}}         &{{21.46}}           &{{24.67}}          &{{27.53}}          &{{11.69}}          &{{14.07}}          &{{15.98}}      \\ \hline

\rowcolor{tels}
FastXML~\cite{prabhu2014fastxml} &{{26.62}}         &{{34.16}}         &{{38.96}}         &{{9.80}}          &{{10.17}}          &{{10.54}}          &{{48.31}}           &{{60.26}}           &{{69.30}}           &{{22.50}}          &{{21.80}}          &{{22.40}}         &{{19.37}}           &{{23.26}}          &{{26.85}}          &{{9.77}}          &{{11.69}}          &{{13.25}}   \\ 

\rowcolor{newton}
PfastreXML~\cite{jain2016extreme}  &{{43.86}}         &{{45.72}}         &{{46.97}}         &{{19.02}}          &{{18.34}}          &{{18.43}}          &{{64.65}}           &{{72.36}}           &{{75.48}}           &{{29.20}}          &{{27.60}}          &{{27.70}}         &{{29.30}}           &{{30.80}}          &{{32.43}}          &{{18.81}}         &{{22.14}}          &{{23.67}}   \\ 

\rowcolor{tels}
Parabel~\cite{prabhu2018parabel} &{{36.36}}         &{{44.04}}         &{{48.29}}         &{{11.66}}          &{{12.73}}          &{{13.68}}          &{{50.93}}           &{{64.00}}           &{{72.08}}           &{{28.80}}          &{{31.90}}          &{{34.60}}         &{{25.43}}           &{{29.43}}          &{{32.85}}          &{{12.82}}          &{{15.61}}          &{{17.73}}       \\ 

\rowcolor{newton}
Bonsai~\cite{khandagale2019bonsai}  &{{37.33}}         &{{45.40}}         &{{49.92}}         &{{11.85}}          &{{13.44}}          &{{14.75}}          &{{51.30}}           &{{64.60}}           &{{72.48}}       &{{27.84}}           &{{32.75}}          &{{36.26}}    &{{27.08}}          &{{30.79}}          &{{34.11}}       &{{13.79}}          &{{16.71}}          &{{18.87}}   \\ \hline

\rowcolor{tels}
SLICE~\cite{jain2019slice}    &{{42.12}}         &{{44.28}}         &{{48.69}}         &{{12.24}}          &{{13.98}}          &{{14.89}}          &{{50.69}}           &{{64.38}}           &{{71.64}}           &{{29.04}}          &{{32.98}}          &{{36.86}}         &{{28.14}}           &{{31.14}}          &{{34.33}}          &{{14.01}}          &{{16.98}}          &{{19.01}}    \\  

\rowcolor{newton}
PD-Sparse~\cite{yen2016pd}  &{{38.28}}         &{{42.00}}         &{{44.89}}         &{{10.78}}          &{{11.42}}          &{{12.24}}          &{{49.58}}           &{{61.63}}           &{{68.23}}           &{{--}}          &{{--}}          &{{--}}         &{{--}}           &{{--}}          &{{--}}          &{{--}}   &{{--}}          &{{--}}  \\ 

\rowcolor{tels}
PPDSparse~\cite{yen2017ppdsparse}    &{{\underline{37.61}}}         &{{\underline{46.05}}}         &{{\underline{50.79}}}         &{{12.84}}          &{{13.62}}          &{{14.98}}          &{{51.62}}           &{{65.24}}           &{{70.71}}           &{{31.48}}          &{{33.86}}          &{{38.21}}         &{{\underline{26.64}}}           &{{\underline{30.65}}}          &{{\underline{34.65}}}          &{{13.12}}          &{{15.89}}          &{{18.16}}     \\

\rowcolor{newton}
DiSMEC~\cite{babbar2017dismec}    &{{\underline{41.20}}}         &{{\underline{45.40}}}         &{{\underline{49.30}}}         &{{\underline{13.60}}}          &{{\underline{13.10}}}          &{{\underline{13.80}}}          &{{\underline{59.10}}}           &{{\underline{67.10}}}           &{{\underline{71.20}}}           &{{\underline{31.20}}}          &{{\underline{33.40}}}          &{{\underline{37.00}}}         &{{\underline{27.80}}}           &{{\underline{30.60}}}          &{{\underline{34.20}}}          &{{12.46}}          &{{14.49}}          &{{17.48}}  \\ 

\rowcolor{tels}
ProXML~\cite{babbar2019data}    &{{45.20}}         &{{48.50}}         &{{51.00}}         &{{14.27}}          &{{15.91}}          &{{16.97}}          &{{62.66}}           &{{66.28}}           &{{74.16}}           &{{33.10}}          &{{35.20}}          &{{39.40}}         &{{30.80}}           &{{32.80}}          &{{35.10}}          &{{11.23}}          &{{13.69}}          &{{15.76}}   \\ \hline

\rowcolor{newton}
XML-CNN~\cite{liu2017deep}  &{{32.41}}         &{{36.95}}         &{{39.45}}         &{{9.39}}          &{{10.00}}          &{{10.20}}          &{{52.42}}           &{{62.83}}           &{{67.10}}        &{{29.48}}           &{{32.68}}          &{{36.37}}    &{{17.43}}          &{{21.66}}          &{{24.42}}                  &{{--}}          &{{--}}          &{{--}}  \\ 

\rowcolor{tels}
AttentionXML~\cite{attentionxml}    &{{44.97}}         &{{51.91}}         &{{54.86}}         &{{15.57}}          &{{16.80}}          &{{17.82}}          &{{53.76}}           &{{68.72}}           &{{76.38}}           &{{30.29}}          &{{33.85}}          &{{37.13}}         &{{30.85}}           &{{38.97}}          &{{44.12}}          &{{15.52}}          &{{18.45}}          &{{20.60}}     \\

\rowcolor{newton}
X-BERT~~\cite{chang2019x}     &{{43.21}}         &{{48.24}}         &{{50.15}}         &{{13.68}}          &{{14.52}}          &{{15.96}}          &{{52.98}}           &{{66.97}}           &{{74.29}}           &{{30.12}}          &{{32.79}}          &{{36.69}}         &{{31.33}}           &{{39.60}}          &{{45.61}}          &{{16.68}}          &{{19.97}}          &{{22.16}}    \\

GNN-XML &\textbf{47.84}         &\textbf{52.95}         &\textbf{55.64}         &\textbf{17.48}          &\textbf{18.34}         &\textbf{19.95}          &\textbf{69.52}     &\textbf{73.12}          &\textbf{77.76}          &\textbf{32.27}          &\textbf{35.63}         &\textbf{40.15}         &\textbf{35.46}           &\textbf{42.05}          &\textbf{47.37}          &\textbf{21.38}          &\textbf{23.22}          &\textbf{24.52}    \\ \bottomrule
\end{tabular}}
\caption{P@$k$ and S@$k$ (shorthand for PSP@$k$) on benchmark datasets for $k=1,3,5$. The best score for each metric is highlighted in bold. Entries marked "--" imply the scores are unavailable due to memory limitation. Underlined entries indicate results directly from the Extreme Repository.}
\label{main_results}
\end{table}

\subsection{Comparisons with State-of-the-Arts} 
\label{ssub:results}
Table~\ref{main_results} shows the performance results w.r.t P@$k$ and PSP@$k$ among the competing methods. We also evaluated the performance by nDCG@$k$ and PSnDCG@$k$, and confirmed that their performance kept the same trends as P@$k$ and PSP@$k$, respectively. We thus omit the results due to space limitation (see Sup.~{\color{red}{B}}). Overall, GNN-XML consistently achieves the highest scores on all datasets across four evaluation metrics. Concretely, GNN-XML shows consistent improvement over previous state-of-the-art method X-BERT on all datasets in terms of P@1, P@3 and P@5, validating that our model has more powerful text representational capabilities and a better generalization ability. An important observation should be noted: GNN-XML outperforms all compared approaches by a large margin on PSP@$k$ with $k=1,3,5$, that is, the metrics for evaluating performance on \textit{tail labels}, revealing the great advantages on recognition of \textit{tail labels} over all competitors. This significant gain stems from two novel techniques, the bilateral-branch graph isomorphism networks with more fine-grained text modeling and the formation of dependency-aware and semantic-aware label clusters. The former learns the universal patterns and shifts the learning attention to tail data gradually, implying that our model can learn matching signals to clusters that have diverse tail labels and small number of relevant training instances. The latter sustains the diversity in the label space by not enforcing any constraints of partitions, thus enabling tail labels to be separate clusters if they are indeed  different from others.


\subsection{Ablation Studies} 
\label{sub:ablation_studies}

\begin{figure}[!t]
\begin{tabular}{@{}c@{}}
\includegraphics[width=1.0\textwidth]{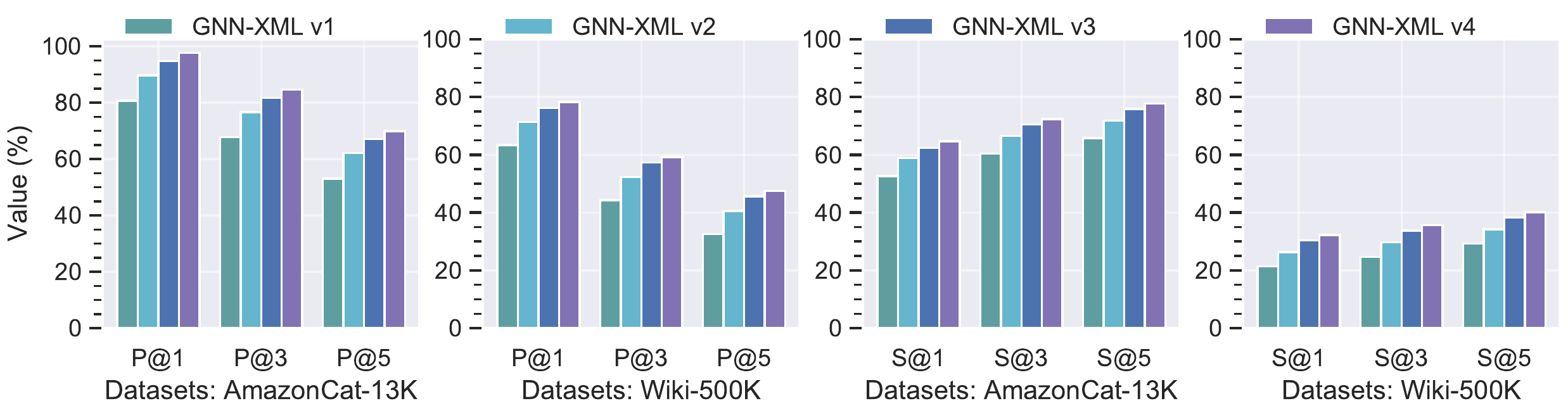}
\end{tabular}
\caption{Ablation study of the graph matching model on AmazonCat-13K and Wiki-500K.}
\label{graph-matching}
\end{figure}

\begin{figure}[!t]
\centering
\begin{tabular}{@{}c@{}}
\includegraphics[width=1.0\textwidth]{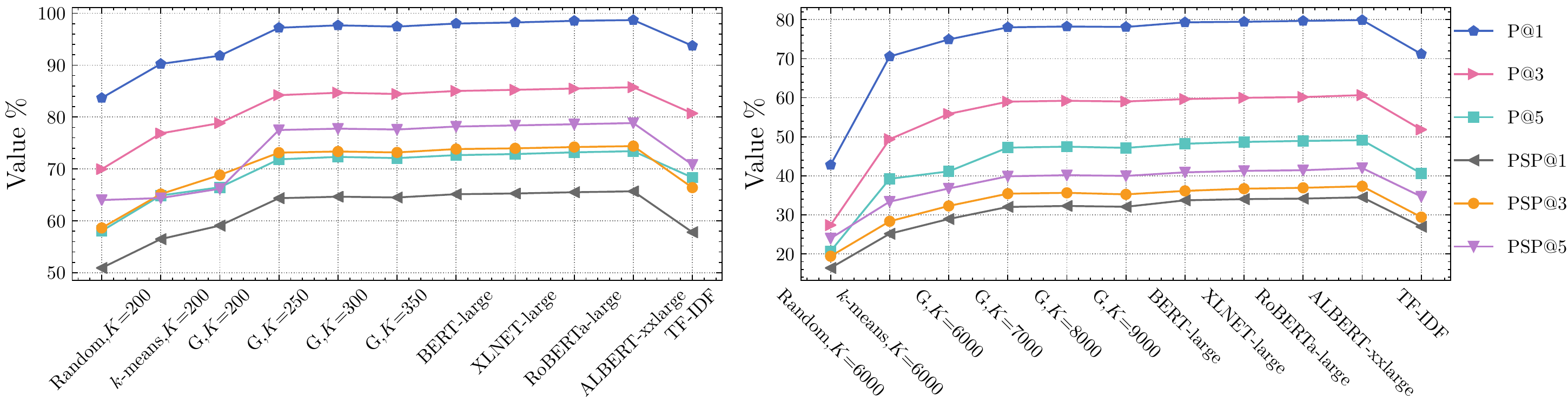}
\end{tabular}
\caption{Ablation study on the AmazonCat-13K (\textit{right}) and Wiki-500K (\textit{left}) datasets.}
\label{label-clustering}
\end{figure}

We empirically examine the impact of main components of GNN-XML via ablation tests, as illustrated in Fig.~\ref{graph-matching} and Fig.~\ref{label-clustering}. We show the effects of three components on the graph matching model in Fig.~\ref{graph-matching}. Overall, we can see that each of the three components consistently boosted the performance of GNN-XML. We first simply use the sum aggregation of all vertex embeddings as the text representation (v1). The insertion of a single conventional branch with GIN leads to performance improvement (v2), indicating a more powerful representational capacity resulting from GIN. Then the addition of re-balance branch further improves the performance (v3), validating the effectiveness of the bilateral-branch learning in alleviating tail-label problems. Finally, the READOUT also results in performance gain (v4). This suggests that integrating multi-scale structure information is favorable for XMTC. Fig.~\ref{label-clustering} shows the influence of three components on the label clustering model, including label partitioning approaches, number of clusters and label representations. We can observe that random partition (Random) tends to yield poor performance. This is because while random partition also produces diverse clusters (see Fig.~\ref{long-tail}), it fails to form meaningful clusters, which poses difficulties in the matching. Our label partitioning (G) consistently outperforms $k$-means, implying that clusters with more semantic awareness and dependency awareness are constructed, which facilitate the matching. Initially increasing the number of clusters $K$ from small to large will improve performance. However, when $K$ reaches a certain range, increasing $K$ further leads to a slight performance improvement, and the optimal performance occurs when K reaches around $\left \lfloor L/60 \right \rfloor$.  Additionally, we also use some other advanced pre-trained language embeddings as well as the TF-IDF features for label representations. Quantitative results show that using alternatives of BERT-base, such as XLNET, RoBERTa and ALBERT, will further lead to performance gains. However, considerable performance degradation occurs when substituting BERT-base with TF-IDF, which gives strong evidence of the superiority of dense features over sparse BOW features.


\paragraph{Training/Prediction Time and Model Size} 
\label{par:model_size_and_training_prediction_time}
We detailedly record the model size and training time of GNN-XML and its competitors on all datasets (refer to Sup.~{\color{red}{B}}). We observe empirically that GNN-XML achieves the best performance at the expense of a slightly larger training time and model size compared to X-BERT. It may be noted that the training phase can be conducted in an off-line fashion, and GNN-XML can be easily trained with limited memory space of GPU (16GB) and RAM (100G) via efficient graph sampling strategy. Crucially, the prediction time of GNN-XML is in the order of milliseconds. Hence it meets the low latency requirements of real-time XMTC applications.




\section{Conclusion} 
\label{sec:conclusion}
In this paper, we present GNN-XML, an efficient and effective graph neural network framework for solving the XMTC problem. At a high level, we achieve the \textit{scalability} by partitioning an enormous number of labels into a small number of clusters with a novel label graph partitioning  method, and alleviate the data \textit{sparsity} by developing a powerful bilateral-branch graph isomorphism network to decouple conventional representation learning and classifier learning for tail labels. Numerical results demonstrate that GNN-XML is significantly better than state-of-the-arts while maintaining comparable prediction efficiency and model size.

\bibliographystyle{hplain}
\bibliography{main}

\end{document}